\documentclass[letterpaper, 10 pt, conference]{ieeeconf}  
\IEEEoverridecommandlockouts                              
\usepackage{amssymb}
\usepackage[table,xcdraw]{xcolor}
\usepackage[most]{tcolorbox}
\usepackage{bm}
\usepackage{amsmath}
\usepackage{amsfonts}

\DeclareMathAlphabet{\mathpzc}{OT1}{pzc}{m}{it}

\DeclareFontFamily{U}{jkpmia}{}
\DeclareFontShape{U}{jkpmia}{m}{it}{<->s*jkpmia}{}
\DeclareFontShape{U}{jkpmia}{bx}{it}{<->s*jkpbmia}{}
\DeclareMathAlphabet{\mathfrak}{U}{jkpmia}{m}{it}

\DeclareMathOperator*{\argmin}{arg\,min}

\newcommand{\R}{\mathbb{R}}

\newcommand{\N}{\mathcal{N}}

\renewcommand{\L}{\mathcal{L}}

\newcommand{\M}{\mathcal{M}}

\definecolor{black}{rgb}{0.3,0.3,0.3}
\definecolor{blackb}{rgb}{0.3,0.3,0.3}
\definecolor{brownb}{rgb}{0.3,0.3,0.3}

\newtcbtheorem[no counter]{Definition}{Definition}{
  enhanced,
  rounded corners,
  attach boxed title to top left={
    yshifttext=-1mm
  },
  colback=white,
  colframe=black,
  fonttitle=\bfseries,
  coltitle=white,
  boxrule=0.5pt,
  boxed title style={
    rounded corners,
    size=small,
    colback=black,
    colframe=black,
  } 
}{def}

\usepackage{subfig}
\usepackage{booktabs}
\usepackage{placeins}
\title{\LARGE \bf
Multi-Embodiment Robotic Retargeting via Guided Diffusion Model 
}

\author{
Zhefeng Cao$^{1,2}$, Ben Liu$^{1}$, Shunpeng Yang$^{1,2}$,
Sen Li$^{2}$, Wei Zhang$^{1,4}$, Hua Chen$^{3,4}$%
\thanks{$^1$ Southern University of Science and Technology, Shenzhen, China. (corresponding author: zhangw3@sustech.edu.cn)}
\thanks{$^2$ The Hong Kong University of Science and Technology, HongKong, China}
\thanks{$^3$ Zhejiang University-University of Illinois Urbana-Champaign Institute, Haining, China}
\thanks{$^4$LimX Dynamics, Shenzhen, China.}
}

\usepackage{url}

\begin{document}

\maketitle
\thispagestyle{empty}
\pagestyle{empty}

\begin{abstract}
The retargeting of diverse and kinematically feasible reference trajectories from human demonstrations facilitates the transfer of human skills to robots, thereby simplifying the learning of complex tasks. However, existing motion retargeting pipelines are heavily tied to specific skeletons with individually hand-crafted settings, which limits scalability across heterogeneous embodiments. 
In this work, our unified framework transfers given reference motions to multiple heterogeneous robots via a transformer-based diffusion model. We represent robotic skeletons as graph structures, enabling effective extraction of topological and geometrical properties via customized attention mechanisms. Due to the scarcity of high-quality motion datasets for diverse robots, we guide the diffusion model with energy-based retargeting losses.
Experiments across heterogeneous robots validate the effectiveness of our unified kinematically feasible retargeting framework on controllable kinematic structures, unseen embodiments, diverse motion domains, and sim-to-sim whole-body tracking tasks.


\end{abstract}

\section{INTRODUCTION}

Legged robots have demonstrated impressive dynamic capabilities~\cite{van24,li24,rad24} such as walking, running, and jumping, thanks to advancements in reinforcement learning. However, each motion still requires independently tuned reward functions~\cite{van24, jiang24}, which hinders long-horizon or complex tasks. To alleviate this, mainstream studies~\cite{he24h2o,he24omnih2o,ji24exbody2} incorporate imitation learning to utilize kinematically feasible joint-level motion references, which are retargeted from human or other robot demonstrations.
However, existing retargeting methods~\cite{ayusawa17,yoon24,he24h2o} are typically tailored to specific robots, limiting scalability and requiring manual adjustments that do not fundamentally resolve the overhead. In this work, we focus on developing a unified kinematically feasible motion retargeting framework for heterogeneous robotic embodiments to reduce the need for robot-specific engineering.

Generalizing existing retargeting pipelines to a group of embodiments faces the challenge of kinematically inconsistent data structure, both optimization-based methods~\cite{hu14,gomes19,mink} and learning-based methods~\cite{choi20,yan23} require adapting the solver configurations or network architectures for different embodiment pairs. In the field of animation, the work in \cite{gat25} generates motions for various animal characters leveraging skeleton information, which can differ in joint numbers and even kinematic chains. This method illustrates the potential of using skeletal structures to generalize motion generation across diverse morphologies. However,  methods in animation only consider the characters represented with uniform kinematic structures and unrestricted three-degree-of-freedom (DOF) ball joints, which widely differ with real robotic embodiments whose single-degree-of-freedom actuated joints operate under strict physical and kinematic limits. Furthermore, unlike human characters that benefit from large-scale motion datasets~\cite{AMASS, LAFAN1, HumanML}, high-quality motion data for different robotic embodiments are largely unavailable.
\begin{figure}[t]
    \centering
    \includegraphics[width=0.88\linewidth]{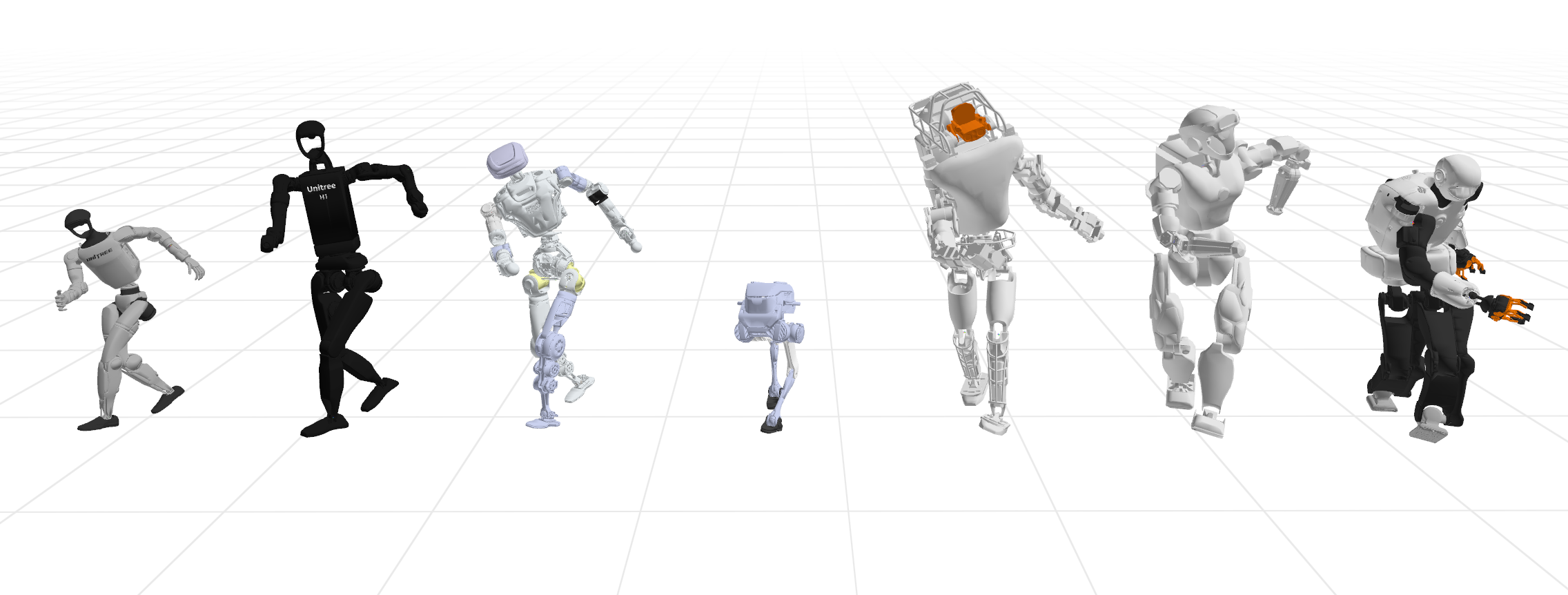}
    \caption{\textbf{Multi-embodiment motion retargeting.} Our unified framework enables the transfer of a single motion to multiple heterogeneous embodiments via a guided diffusion model.}
    \label{fig:fr}
    \vspace{-6pt}
\end{figure}

To address these challenges and enable a unified approach to motion retargeting across diverse, heterogeneous embodiments, we propose a transformer-based diffusion framework that leverages reference motions as conditional guidance for kinematically feasible motion retargeting across multiple embodiments. Our framework is conditioned on three components: the reference motion, the skeletal information of the reference embodiment, and that of the target embodiment. The reference motion is encoded by a Transformer decoder, which are capable of handling variable-length tokens and incorporating graph-structured skeleton condition through attention mechanisms. The denoising network architecture is based on a similar Transformer decoder absorbing the information of both reference motion and target skeleton with graph representation. Besides, the joint limit constraints are enforced by a clip network block. To mitigate the scarcity of robotic motion datasets, our diffusion model is trained with energy-based retargeting guidance computed from the reference motion, which provides a kinematics-aware supervisory signal. 

We conduct a series of comprehensive experiments to validate the effectiveness of our framework across multiple robotic embodiments. The qualitative results and corresponding analyses highlight the model’s capability to capture graph-structured features and its generalization to unseen skeletons and diverse motion patterns. We further apply the generated motions of two non-homeomorphic robots to motion mimic tasks with successful sim-to-sim deployment. Through ablation studies, we further analyze the contributions of different components in our framework.

In summary, our contributions are as following:
\begin{itemize}
    \item We propose a unified transformer-based guided diffusion framework for kinematically feasible multi-embodiment robotic motion retargeting.
    \item The coordination of the attention mechanisms and the graph-based skeleton representation enables the diffusion model to capture and interpret skeletal differences among distinct embodiments.
    \item The proposed energy-based retargeting guidance from reference motions, effectively addresses the lack of high-quality motion datasets for diverse robotic embodiments.
    \item Our experiments validate the effectiveness of the proposed framework for multiple embodiments and demonstrate its successful application to long-horizon robotic whole-body tracking, verifying its practical applicability.
\end{itemize}

\section{Related Works}

\subsection{Multi-embodiment Motion Retargeting}
Traditional motion retargeting methods~\cite{hu14, gomes19,ayusawa17} rely on the construction of constrained optimization with hand-crafted kinematic objectives, such as tracking keypoint positions and link orientations, which only apply to specific motions and embodiments. 
Most learning works~\cite{choi20, aberman20, yan23} in retargeting also consider a single embodiment, whose network structures cannot fit the configuration of other embodiments or require separate training for them.
Only a few motion generation works~\cite{yang24, tripathi24} in animation explore multiple embodiments in a unified manner, where \cite{tripathi24} uses a conditional VAE framework to generate motions based on SMPL model~\cite{loper23} with different body shapes. Recently, \cite{gat25} adopts a diffusion-based framework to generate motions for heterogeneous animal characters. However, since no conditioning is provided, the model cannot be used for retargeting tasks directly. Moreover, for robotic embodiments, additional challenges arise from strict joint-limit constraints, and the lack of open-source motion datasets for diverse robots further complicates the problem.


\subsection{Graph-based Representation}
Early work on capturing the graph features of different embodiments is based on Graph Neural Networks~\cite{wang18,huang20} with a message passing mechanism. However, message passing can only aggregate information from neighboring nodes, which mainly captures local features. Although GraphSAGE~\cite{hamilton17} is proposed to consider the k-nearest neighbors, it is hard to learn a global latent graph representation for downstream tasks. Then transformer-based architectures are proposed to extend the restricted attention over all nodes by the attention mechanism. Body Transformer~\cite{sfer24} uses the attention mask to add the connectivity information of the robot skeleton, where it can only aggregate information from neighboring joints and does not consider other properties of the skeleton graph (bone offsets, joint rotation axis) into the Transformer. \cite{gat25} captures both spatial and temporal features of the training motion by a spatial-temporal attention block, where the spatial relationship between each joint is enriched by incorporating the graph information with GRPE mechanism~\cite{park22}. Inspired by their work, we incorporate different components of the graph data with different attention mechanisms, including the discrete joint correspondence information.

\subsection{Diffusion Model with Guidance}
Diffusion models have demonstrated impressive performance across a range of generative tasks \cite{ho2020DDPM,song2019gradient,song2020SDE,karras2022EDM}. While standard diffusion models are typically designed to learn and sample from an unconditional data distribution, many practical applications require sampling from a conditional distribution. To enable conditional generation, classifier-guided \cite{dhariwal2021ClassifierGuided} and classifier-free guided \cite{ho2022classifierfreeguided} approaches have been proposed. These methods aim to learn a conditional score function by incorporating additional information during sampling. However, they rely on either an external classifier or access to paired conditional data, which poses significant challenges in motion retargeting scenarios where such data is scarce. An alternative line of work introduces energy-based guidance by modifying the sampling distribution to $p(x)\propto p_\texttt{uncond}(x)\exp(-\mathcal{E}(x))$ \cite{janner2022planningflexible,lu2023contrastive}, 
where the energy function $\mathcal E(x)$ encodes task-specific preferences or performance criteria. This approach is particularly appealing for retargeting tasks, as it allows explicit modeling of desired motion characteristics without requiring paired data.

\section{Guided Diffusion for Multi-embodiments Retargeting}
In this section, we first introduce the motion representation and the graph-based construction of unified robotic skeletons, which serve as the foundation for handling diverse robotic embodiments. We then present network structures of the reference motion encoder and the diffusion denoising process, detailing the architecture and the design of each internal attention block. Finally, we describe the diffusion process with retargeting guidance and formulate the corresponding training objectives.

\begin{figure*}[t!]
    \centering
    \includegraphics[width=0.80\linewidth]{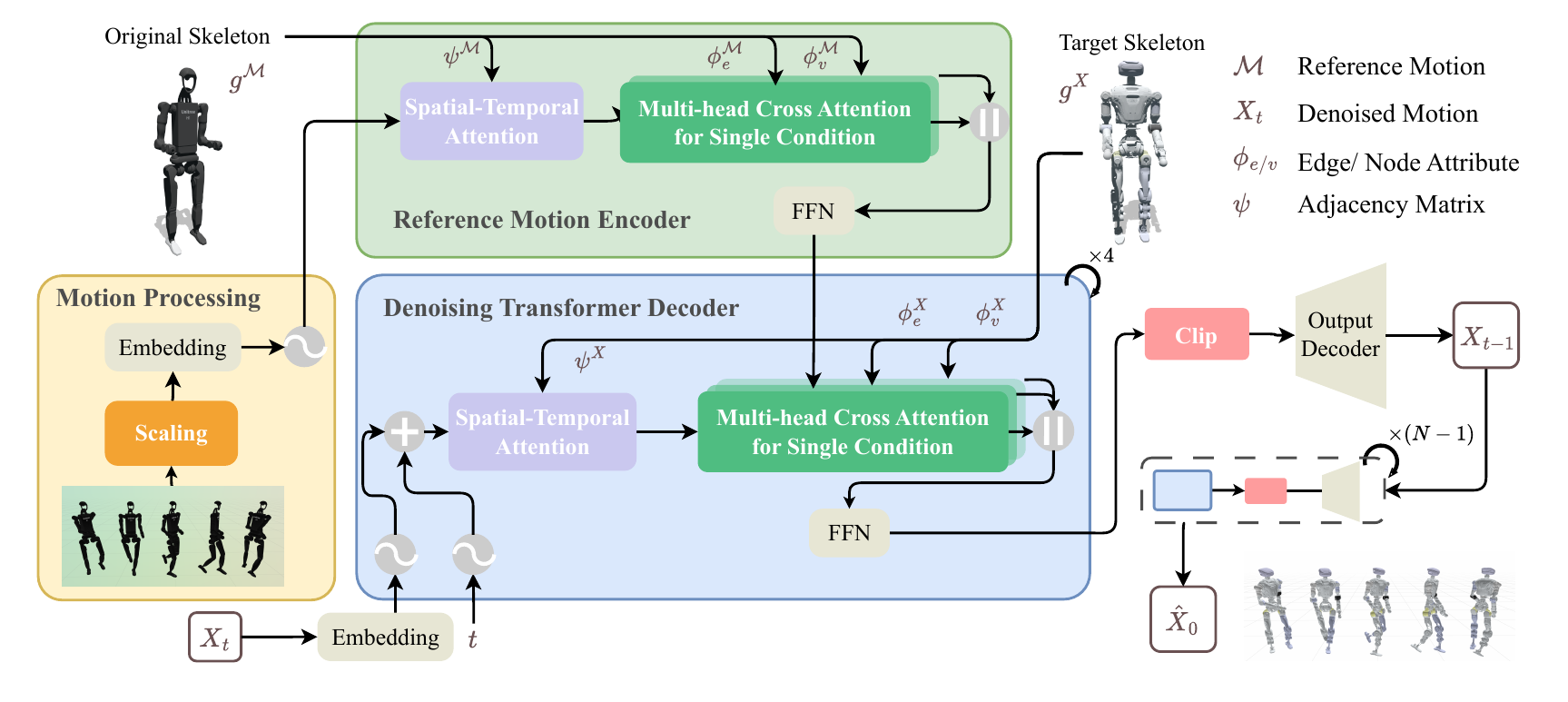}
    \vspace{-10pt}
    \caption{\textbf{Overview of our framework}. The denoising network is based on a transformer decoder with a noisy motion input $X_t$ under the conditions of the reference motion $\M$ and graphs $g^{\M,X} = \{ \phi_v^{\M,X}, \phi_e^{\M,X},\psi^{\M,X}\}$. The input motion is tokenized at the joint level, where the base joint and other joints are embedded by independent encoders. Then spatial and temporal attentions extract the relationships between all joints and the chronological relationships along the time window. In addition, the spatial attention absorbs the joint connectivity $\psi$ to enrich the joint relationships. For other graph conditions, we use a multi-conditional cross attention to treat them individually. In particular, the reference motion condition $\M$ is encoded using a similar transformer decoder and incorporates the joint correspondence $\eta$ as an attention mask.}
    \label{fig:framework}
    \vspace{-15pt}
\end{figure*}

\subsection{Data Representation}
\subsubsection{Motion representation} 
The motion sequence of robots is denoted by a 3D tensor $ \M \in \R^{T\times J \times D}$, where $ T $ is the number of motion frames, $ J $ is the number of joints including the floating base, and $D$ is the dimension of each joint features. Without loss of generality, we assume that all robot joints, except for the base joint, are one-DOF revolute joints. In this way, all joints have the same semantic motion features, which can be encoded by the same encoder later. By adopting redundant joint features inspired by ~\cite{gat25, guo22}, we denote the j-th joint (except the base joint) features in i-th frame as $\M_j^i = [q_j,p_j,v_j] \in \R^7$ where $q_j \in \R$ is 1D joint rotation angle, $p_j \in \R^3$ is 3D global joint position and $v_j \in \R^3$ is linear velocity. For the base joint, its features are denoted as $ \M_0^i = [r_0,p_0,v_0] \in \R^9$ where $r_0 \in \R^3$ is the base orientation represented by axis-angle, $p_0 \in \R^3$ is 3D base position, and $v_0 \in \R^3$ is linear velocity. To keep a unified data structure, we pad the features of other joints to $ D = 9$, matching the dimension of the base joint.

\subsubsection{Graph-based skeleton representation} 
We propose to utilize the directed acyclic graph (DAG) $g\in \mathcal{G}=\{ \phi_v, \phi_e,\psi\}$ to represent the skeleton structure of the robot, which can be viewed as a specific way of interpreting the kinematic tree of the robot.
The edge attribute $\phi_e$ and node attribute $\phi_v$ represent the robot's configuration at the initial pose, with zero joint rotations. In detail, the node attribute $\phi_v\in R^{3}$ is a 3D unit vector representing the joint rotation axis, and the edge attribute $\phi_e\in R^{3}$ is a 3D vector pointing from parent joint to child joint whose norm represents the link length. The connectivity of the graph is represented by an adjacency matrix $\psi\in \N^{J\times J}$, where the integral elements $\{0,1,2,3\}$ indicate four types of relationships between the joints indexed by the rows and columns: \textit{no relation}, \textit{self}, \textit{parent}, and \textit{child}. This carefully designed representation allows the network to capture the precise graph structure, enabling it to generate motions for desired skeletons effectively.
In addition, to accurately extract the information of the reference motion, we specify the joint correspondence between two embodiments by constructing a joint map $\eta \in \R^{J \times J}$, where the element $\eta_{ij} = 1$ clarifies that the i-th joint of the first embodiment corresponds to the j-th joint of the second embodiment. In this work, $\eta$ is manually specified from semantic key joints.


\subsection{Network Architecture}
Due to the heterogeneous kinematic structures across embodiments, we adopt a transformer-based denoising network to handle structural variability. Specifically, we decompose the motions at the joint level by encoding each joint information in each frame as separate token. So, for each motion there will be $T\times J$ tokens. To purely learn the intrinsic spatial and temporal relationship of the motion, we apply positional encoding preserving the order of the whole input motion tokens, and adopt the transformer decoder to treat the conditions independently by using cross-attention. At each denoising step $t$, the denoising network $f_{\theta}(X_t|\M,g^{\M},g^{X})$ predicts the desired retargeted motion $\hat{X}_0$ from a noisy motion $X_t$, given the reference motion $\M$ with the matching skeleton $g^{\M}$ and desired skeleton $g^{X}$.

The whole network architecture consists of two transformer decoders shown in Figure~\ref{fig:framework}, one is used for encoding the reference motion, another is for predicting retargeted motion. Each transformer decoder has a similar structure but with different conditions, including the spatial-temporal attention block and multi-conditional cross attention block.

\paragraph{Spatial-temporal attention block}
The spatial-temporal attention block models all joint dependencies within each frame and temporal correlations across frames. Given motion tokens $X_{\text{in}} \in \R^{T\times J \times H}$, where $H$ is the latent dimension, we apply spatial attention over joints and temporal attention over frames in a sequential manner, where both follow the same attention formulation below with subtle differences ($A$ for spatial attention and $M_T$ for temporal attention). For an input embedding $X_{\text{in}}$, the attention update is expressed as
\begin{equation}
    X_{\text{out}} = \text{LayerNorm}\Big(X_{\text{in}} + \text{Dropout}\big(\sigma(\tfrac{QK^{\mathsf{T}} + A}{\sqrt{D_h}} + M_T)V\big)\Big),
\end{equation}
where $Q = X_{\text{in}}W_Q$, $K = X_{\text{in}}W_K$, and $V = X_{\text{in}}W_V$ are queries, keys and values, $\sigma$ is the softmax function. Following the work of GRPE~\cite{park22}, we integrate the discrete joint relationships $\psi$ into the spatial attention module using discrete embeddings for queries and keys, denoted as $E^R_q, E^R_k \in \R^{4\times H}$. Then the additional attention score $A$ is constructed by these embeddings as $A_{ij} = q_i \cdot E^R_q [\psi_{ij}] + k_j \cdot E^R_k[\psi_{ij}]$, where $q_i, k_j$ represent the i-th query and j-th key respectively, and $[\cdot]$ means the indexing operation. Additionally, to simplify computation and improve robustness to temporal noise, we apply a temporal window with horizon $T_w$ as an attention mask $M_T$ for the temporal attention module.

\paragraph{Multi-conditional cross attention block}
Besides the connectivity information from the skeleton graph, we expect our model to recognize the node attributes and edge attributes for different embodiments, i.e. joint rotation axes, 3D link vectors, and joint limits, in order to generate motions that respect the corresponding kinematic constraints. Because different conditions have different semantic contents, it is not appropriate to sum all condition embeddings and apply the same parameters to them. To handle multiple conditions simultaneously, we allocate the attention heads equally to each condition and then concatenate all the heads, with each head having its own set of parameters.
\begin{equation}
\begin{aligned}
    &\text{MultiHead}(Q, \{K_{c}, V_{c}\}) = \\
    &\text{Concat}\left( \left\{ \sigma\left( \frac{(Q W^i_Q) (K_{c} W^{i}_K)^T}{\sqrt{D_h}} \right) (V_{c} W^{i}_V) \right\}_{i=1}^{h} \right) W_O,
\end{aligned}
\end{equation}
where $K_{c} = V_{c}\in \R^{N_c\times d}$ are the condition embeddings. In addition, the denoising transformer decoder receives the reference motion embeddings as part of the conditions, which will directly influence the retargeting results. In this work, we set the similarity criterion as the correspondence of the key joint positions. Thus, the useful information of the reference motion within each condition is related to the corresponding joint map $\eta$. In other words, only corresponding joints can affect each other, so we naturally use the temporally extended joint map $\bar{\eta} \in \R^{(T*J)\times (T*J)}$ as an attention mask,
\begin{equation}
    h_{\text{ref}} =  \sigma\left( \frac{(Q W^{\text{ref}}_Q) (K_{\text{ref}} W^{\text{ref}}_K)^T}{\sqrt{D_h}} + \bar{\eta} \right) (V_{\text{ref}} W^{\text{ref}}_V),
\end{equation}
where $K_{\text{ref}}, V_{\text{ref}}$ are the reference motion embeddings.

\subsection{Diffusion Model with Energy-based Guidance}
Motion retargeting can be formulated as a conditional sampling problem that can be tackled by generative learning models, which generate the robotic motion under the conditions of the specified embodiment and the given reference motion of another embodiment. However, unlike other existing motion generation works~\cite{tevet22, chen23, aberman20}, we do not have samples for the desired distribution. In other words, the motion for a specific skeleton that follows a reference motion from another skeleton is unavailable. To address this, we formulate our generative model as a diffusion model with energy-based guidance:
\begin{equation}
\begin{aligned}
    \min_\theta \; \mathbb{E}_{x \sim p_{\mathrm{ref}}(x)} \; 
    \mathbb{E}_{n \sim \mathcal{N}(0, \sigma^2 I)} 
    \Big[ 
        \| D_\theta(x + n, \sigma, C) - x \|_2^2  \\
        + \lambda f_{\mathrm{kin}}( D_\theta(x + n, \sigma, C) )
    \Big],
\end{aligned}
\end{equation}
where $D_\theta(x,\sigma)$ is the denoiser function parametrized by a neural network and related to the score function with $\nabla_x\log p (x;\sigma)\approx(D_\theta(x,\sigma)-x)/\sigma^2$ \cite{karras2022EDM}, $p_{\texttt{ref}}(x)$ is the distribution of the dataset, and $n$ is the noise characterized by $\sigma$. Here we add an extra input $C=(\M, g^{\M}, g^X, \eta)$ to represent the condition in the generating process. 
The first term in the loss represents the distribution of reference motion, which can be regarded as a prior, where $x$ is the sample from the motion dataset.

The second term represents the energy-based guidance $p(x)\propto \exp(-f_{\texttt{kin}}(x))$. 
More specifically, a smaller $f_{\texttt{kin}}$ value indicates that the target position is closer to the desired motion, which means better retargeting performance, hence should have a higher probability. In detail, the guidance loss is characterized by the kinematics relationship as expressed below,
\begin{equation}
    f_{\texttt{kin}}=\L_{\texttt{kp}}+\L_{\texttt{cst}} + \L_{\texttt{vel}} + \L_{\texttt{norm}}.
\end{equation}
$\L_{\texttt{kp}}$ represents the pose errors of the selected keypoints,
\begin{equation}
    \L_{\texttt{kp}} = \sum\limits_{\{(i,j)|\eta_{ij}=1\}} w_0\| \texttt{FK}(\hat{X}_0, g^X, i) \ominus P(\M_{\alpha}, j)\|^2,
\end{equation}
where $\hat X_0 = D_\theta(x+n,\sigma,C)$ is the generated motion, $\texttt{FK}(\hat{X}_0, g^X, i)$ is the forward kinematics mapping joint configuration to the i-th joint frame pose of the specified skeleton $g^X$, while $P(\M_{\alpha},j)$ directly selects the j-th joint frame pose from the scaled reference motion $\M_{\alpha}$. Given the redundancy of our motion representation, the consistency loss $\mathcal{L}_{\texttt{cst}}$ enforces that the joint-space and task-space representations remain consistent under the forward kinematics constraint,
\begin{equation}
    \L_{\texttt{cst}} = \sum\limits_{i=0}^J  w_1\| \texttt{FK}(\hat{X}_0, g^X, i) \ominus P({\hat{X}_0},i)\|^2.
\end{equation}
Additionally, to maintain the velocity feature, $\L_{\texttt{vel}}$  tracks the keypoints velocities:
\begin{equation}
\begin{aligned}
     \L_{\texttt{vel}} = \sum\limits_{k=1}^T\sum\limits_{\{(i,j)|\eta_{ij}=1\}} w_2\| &(\dot{\texttt{FK}}_k(\hat{X}_0, g^X, i) -\\
     &\dot{P}_k(\M_{\alpha},j)\|^2,
\end{aligned}
\end{equation}
where $\dot{\texttt{FK}}_k$ and $\dot{P}_k$ are computed by difference, $k$ is the index along the time axis. Finally, loss $\L_{\texttt{norm}}$ is a regularization term for the stable learning process:
\begin{equation}
\begin{aligned}
    \L_{\texttt{norm}}= w_3 & \Bigg(  \sum\limits_{i=0}^J \|Q(\hat{X}_0,i) - Q_0(i) \|^2 + \\
    & \sum\limits_{k=1}^T \sum\limits_{i=0}^J \| \dot{\texttt{FK}}_k(\hat{X}_0, g^X,i) \| ^2 \Bigg) ,
\end{aligned}
\end{equation}
where $Q(\hat{X}_0,i)$ selects the i-th joint angle from predicted motion $\hat{X}_0$, $Q_0$ is the nominal configuration.
The same $w_0$--$w_3$ setting is used for all reported robots and motions after balancing validation loss magnitudes.




\section{Experiments}
\label{sec:experiments}
In this section, we evaluate our unified \textbf{kinematically feasible} retargeting framework across \textbf{multiple robotic embodiments}, including qualitative/quantitative results, sim-to-sim mimicry, and attention ablations.

\begin{figure*}[tb!]
    \centering
    \includegraphics[width=0.90\linewidth]{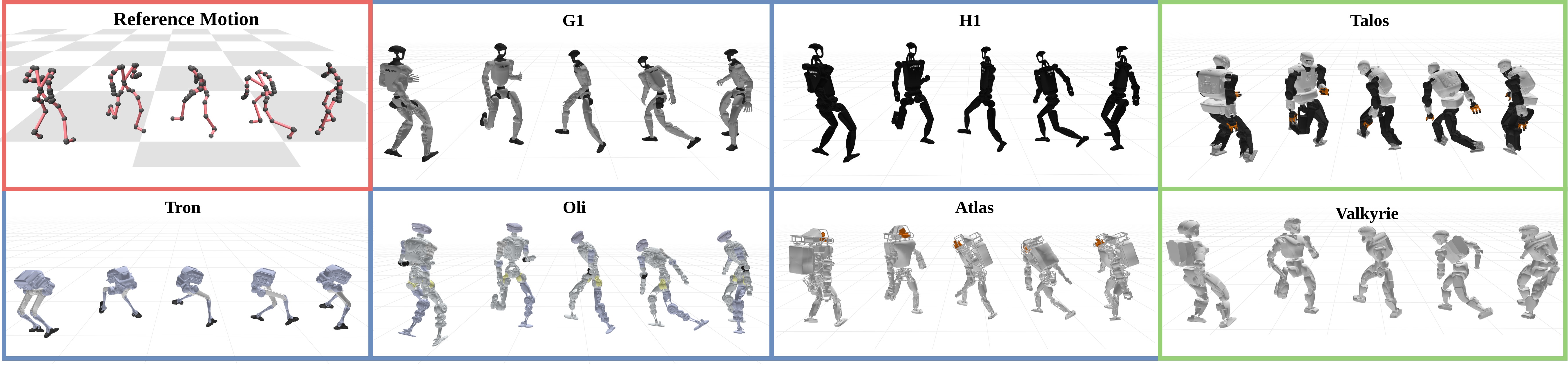}
    \caption{The retargeted motions across multiple robotic embodiments including the new added ones.}
    \label{fig:main_result}
    \vspace{-10pt}
\end{figure*}

\subsection{Experiment Setup}
We train and evaluate on the robotic motion dataset~\cite{lafan1_2025}, which includes the motion sequences of the humanoid robots Unitree G1~\cite{g1} and H1~\cite{H1} mapped from LAFAN1 dataset\cite{LAFAN1}. Each robot in the dataset consists of 40 retargeted motion sequences, totaling over 4 hours, with motion types including `dance', `fallAndGetUp', `fight', `jump', `run', `sprint', and `walk'. We select 3 different motion sequences of each robot from types of `dance', `fallAndGetUp', `fight' as training data, while the remaining data is used as unseen motions for evaluation. All motions are downsampled to 30~Hz and clipped into 60-frame motion clips. To account for the scattered global position distribution after clipping, we set the starting base position of each motion segment to the origin, maintaining a constant height.

For the target robotic embodiments, we directly parse the graph information from the robot URDF files of the Limx humanoid Oli~\cite{Oli}, bipedal robot Tron1~\cite{Tron1} with soles, Atlas~\cite{Atlas}, G1, H1, Talos~\cite{Talos}, and Valkyrie~\cite{valkyrie}. For joint correspondence, we manually identify the indices of the hip, knee, ankle, toe, shoulder, elbow, and hand joints in each skeleton. If any joint is missing in an embodiment, its index is set to -1, facilitating the construction of joint maps for embodiment pairs. Finally, each training sample consists of a reference motion clip, its corresponding skeleton graph, a target skeleton graph and the joint map for retargeting.

\subsection{Implementation Details}
As shown in Fig. \ref{fig:framework}, each reference motion needs to be scaled first to match the size of the target skeleton. We divide the skeleton into several body parts, and the scaling factor $\alpha_k$ of the k-th body part adopts the scaling scheme below, 
\begin{equation}
    \alpha _k= \argmin_{\alpha} \sum_{j\in \mathcal{I}_k} \|\alpha l_j - l_j^o \|^2,
\end{equation}
where $l_j =\sum_{i\in \mathcal{I}_j} \|p_i-p_s\|_2$ and $l_j^o$ are the cumulative body-part link lengths of the target skeleton and original skeleton, $\mathcal{I}_j$ represents the parent index list, $p_s$ is the beginning joint position of the k-th body part. Besides this preprocessing, we also clip the joint angles of the generated motion to ensure the joint-limit constraint during training,
\begin{equation}
    \hat{q}_{\text{clip}}=\sigma(\hat{q})*(\overline{c}^X-\underline{c}^X) + \underline{c}^X,
\end{equation}
where $\sigma$ is the sigmoid function, $\overline{c}^X, \underline{c}^X$ are the joint limits.

To improve adaptability, we augment only the training data with structural perturbations: key links (e.g., thigh and calf) are scaled by factors between 0.5 and 2, other links between 0.67 and 1.5, and certain non-end-effector correspondences are randomly dropped; evaluations use the original URDF skeletons except for controlled tests below.

We use $N=1000$ diffusion steps, latent dimension $H = 240$, batch size $B = 16$, and temporal window size $T_w = 31$. Each Transformer decoder has 4 layers, 6 attention heads, and FFN hidden dimension 1024. Training on one NVIDIA RTX 4090 takes around 18 hours; inference takes about 15 seconds per retargeted motion.

\subsection{Evaluation Metrics}
We evaluate the quality of the retargeted motions from both tracking performance and motion smoothness. Following \cite{xie25}, we use the Global Mean Per Body Position Error ($E_{\text{g-mpbpe}}$, m), Mean Per Body Velocity Error ($E_{\text{mpbve}}$, m/$s$) and Mean Per Body orientation Error ($E_{\text{mpboe}}$, rad) to measure the tracking performance. The motion smoothness is quantified by the average joint-level jerk magnitude ($E_{\text{smooth}}$, rad/$s^3$).

\subsection{Main Results}
To validate our kinematically feasible motion retargeting framework for multiple robotic embodiments, we select five target robots, G1, H1, Atlas, Oli and Tron1. Following \cite{gat25}, there are two distinct types of retargeting skeleton pairs: homeomorphic and non-homeomorphic. The former differ in joint numbers within each kinematic chain but still share the same primal skeleton topology -- all humanoid robots in our study fall into this category. The latter, however, have no common primal skeleton, meaning they differ in body parts and overall structure, such as the pairs between Tron1 and other humanoid robots. In Fig.~\ref{fig:main_result}, we present the retargeted motions for these robots. Our model generates natural and smooth motions by tracking the poses of selected reference joints, including the hip, knee, ankle, toe, shoulder, elbow, and hand. For the biped robot Tron1, which lacks certain joints, the corresponding joint pairs are excluded from the model's input conditions. As a result, Tron1 reproduces only the lower-body portion of the reference humanoid motion. In addition, all generated motions strictly satisfy the joint limit constraints, which make the motions become kinematically feasible. In Table~\ref{tab:baseline}, we further compare our method with the GMR~\cite{GMR} baseline on G1 and H1, the robots supported by the official GMR implementation, under the same in-domain reference motions, where our model achieves comparable performance.
\begin{table}[htb]
\caption{Comparison of retargeting performance against the baseline}
\label{tab:baseline}
\centering
\small
\resizebox{\columnwidth}{!}{%
{\fontsize{8}{8}\selectfont
\begin{tabular}{lllll}
\toprule
\textbf{Methods} & $E_{\text{g-mpbpe}} \downarrow$ & $E_{\text{mpboe}} \downarrow$ & $E_{\text{mpbve}} \downarrow$ & $E_{\text{smooth}} \downarrow$ \\
\midrule
\rowcolor[HTML]{EFEFEF} 
\multicolumn{5}{l}{G1} \\
Ours            & 0.0071\textsubscript{$\pm$0.0038}  & \textbf{0.30\textsubscript{$\pm$0.19}} & \textbf{0.0075\textsubscript{$\pm$0.0035}} & \textbf{0.023\textsubscript{$\pm$0.0073}} \\ 
GMR            & \textbf{0.0056\textsubscript{$\pm$0.017}} & 0.56\textsubscript{$\pm$1.74} & 0.0087\textsubscript{$\pm$0.027} & 0.043\textsubscript{$\pm$0.024} \\ 

\midrule
\rowcolor[HTML]{EFEFEF} 
\multicolumn{5}{l}{H1} \\
Ours          & \textbf{0.012\textsubscript{$\pm$0.0059}} & 0.22\textsubscript{$\pm$0.16} & \textbf{0.013\textsubscript{$\pm$0.0057}} &  \textbf{0.024\textsubscript{$\pm$0.0083}} \\
GMR          & 0.014\textsubscript{$\pm$0.045} & \textbf{0.092\textsubscript{$\pm$0.24}} & 0.013\textsubscript{$\pm$0.036} &  0.048\textsubscript{$\pm$0.026} \\
\bottomrule
\end{tabular}}
}
\end{table}

\begin{table}[htb]
\caption{Retargeting performance across multiple robotic embodiments on in-domain and out-of-domain motions.}
\label{tab:quant-result}
\centering
\small
\resizebox{\columnwidth}{!}{%
{\fontsize{8}{8}\selectfont
\begin{tabular}{lllll}
\toprule
\textbf{Robot} & $E_{\text{g-mpbpe}} \downarrow$ & $E_{\text{mpboe}} \downarrow$ & $E_{\text{mpbve}} \downarrow$ & $E_{\text{smooth}} \downarrow$ \\
\midrule
\rowcolor[HTML]{EFEFEF} 
\multicolumn{5}{l}{In-domain} \\
G1            & \textbf{0.0071\textsubscript{$\pm$0.0038}}  & 0.30\textsubscript{$\pm$0.19} & \textbf{0.0075\textsubscript{$\pm$0.0035}} & 0.023\textsubscript{$\pm$0.0073} \\ 
H1          & 0.012\textsubscript{$\pm$0.0059} & \underline{0.22\textsubscript{$\pm$0.16}} & 0.013\textsubscript{$\pm$0.0057} &  0.024\textsubscript{$\pm$0.0083} \\
Oli           & \underline{0.011\textsubscript{$\pm$0.0052}} & 0.25\textsubscript{$\pm$0.17} & \underline{0.013\textsubscript{$\pm$0.0052}} & 0.023\textsubscript{$\pm$0.0084} \\
Atlas        & 0.016\textsubscript{$\pm$0.0059} & 0.30\textsubscript{$\pm$0.20} & 0.021\textsubscript{$\pm$0.0061} & 0.029\textsubscript{$\pm$0.0084} \\
Tron1        & 0.015\textsubscript{$\pm$0.0054} & 0.57\textsubscript{$\pm$0.27} & 0.029\textsubscript{$\pm$0.0088} & 0.020\textsubscript{$\pm$0.0089} \\
Talos (New)         & 0.014\textsubscript{$\pm$0.0073} & \textbf{0.21\textsubscript{$\pm$0.17}} & 0.015\textsubscript{$\pm$0.0068} &  0.022\textsubscript{$\pm$0.0087}\\
Valkyrie (New)      & 0.015\textsubscript{$\pm$0.0073} & 0.24\textsubscript{$\pm$0.18} & 0.014\textsubscript{$\pm$0.0050} &  0.024\textsubscript{$\pm$0.0097}\\

\midrule
\rowcolor[HTML]{EFEFEF}
\multicolumn{5}{l}{Out-of-domain} \\
G1            & 0.020\textsubscript{$\pm$0.033} & 0.29\textsubscript{$\pm$0.13} & 0.019\textsubscript{$\pm$0.028} &  \underline{0.019\textsubscript{$\pm$0.0064}} \\
H1            & 0.20\textsubscript{$\pm$0.68} & 0.33\textsubscript{$\pm$0.48} & 0.17\textsubscript{$\pm$0.56} & 0.022\textsubscript{$\pm$0.0071} \\
Oli           & 0.12\textsubscript{$\pm$0.39} & 0.32\textsubscript{$\pm$0.35} & 0.12\textsubscript{$\pm$0.40} & 0.020\textsubscript{$\pm$0.0073} \\
Atlas         & 0.65\textsubscript{$\pm$2.02} & 0.47\textsubscript{$\pm$0.75} & 0.59\textsubscript{$\pm$1.82} & 0.027\textsubscript{$\pm$0.011} \\
Tron1         & 0.026\textsubscript{$\pm$0.024} & 0.73\textsubscript{$\pm$0.16} & 0.041\textsubscript{$\pm$0.026} & \textbf{0.017\textsubscript{$\pm$0.0081}} \\
\bottomrule
\end{tabular}}
}
\vspace{-4pt}
\end{table}

\begin{figure}[htb]
    \centering
    \includegraphics[width=0.62\linewidth]{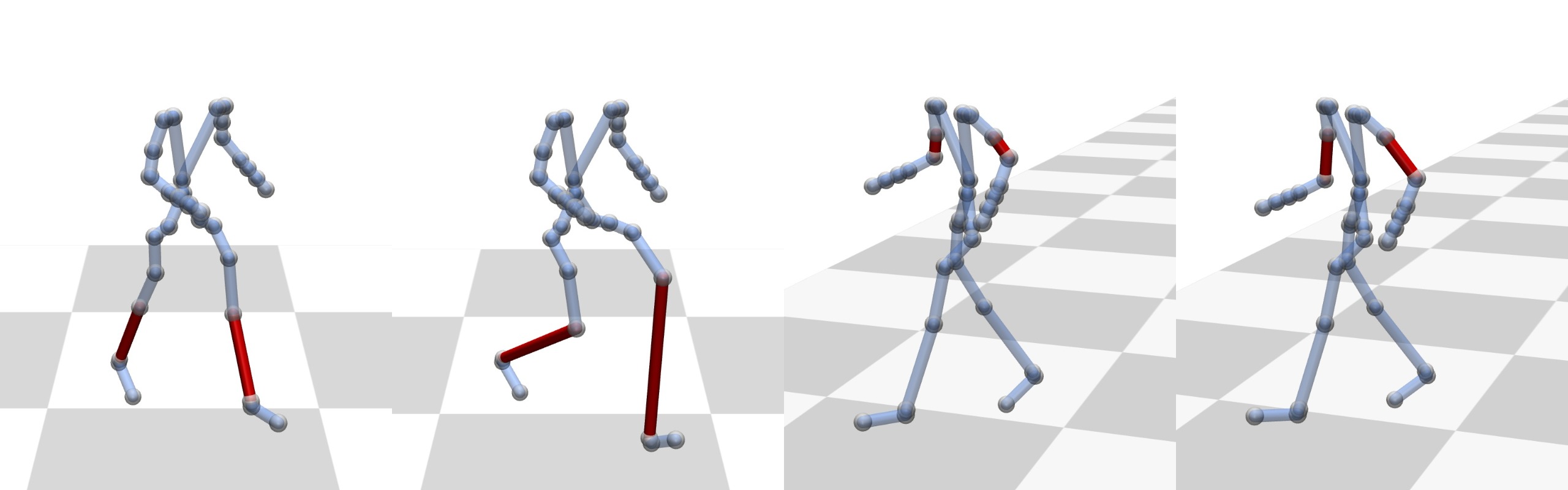}
    \caption{Validation of controllable link-length variations. The red links are the scaled ones with a factor of 2.}
    \label{fig:link-length}
\end{figure}
\begin{figure}[htbp]
    \centering
    \includegraphics[width=0.62\linewidth]{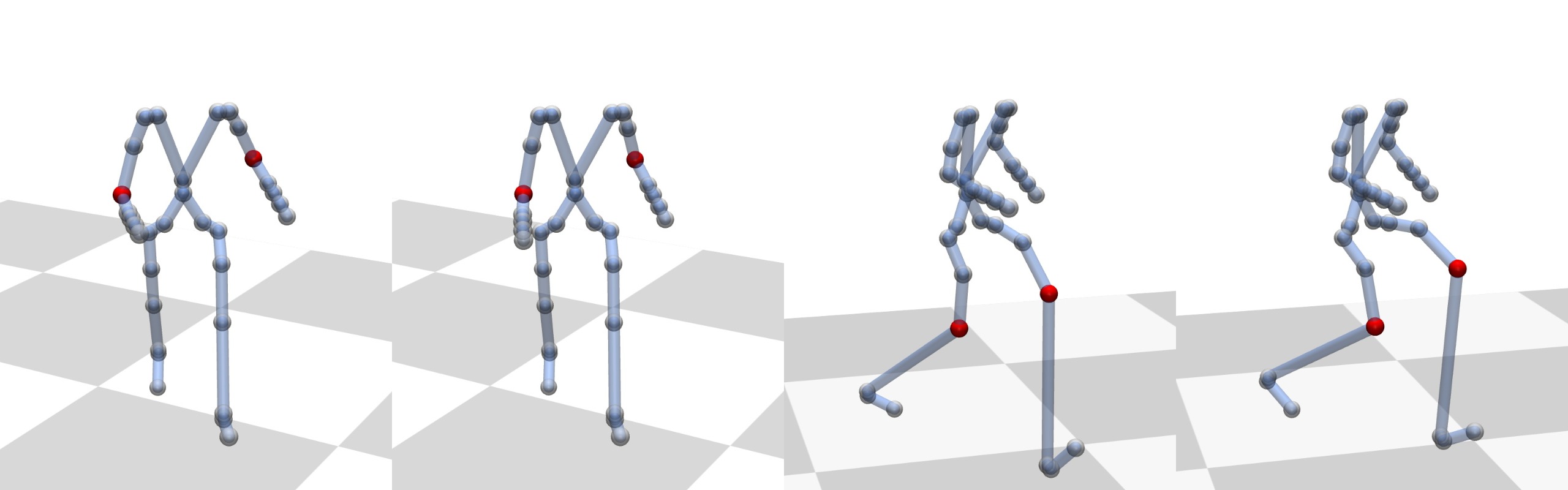}
    \caption{Validation of controllable joint correspondence input. The red joints are the test ones. The first and the third subfigs require the specified red joint correspondence.}
    \label{fig:correspondence}
\end{figure}

To further demonstrate the model's adaptability, we vary link lengths and joint correspondences in the input skeletons. In the link-length experiment, we remove knee or elbow correspondence and change the calf or upper-arm length. As shown in Fig. \ref{fig:link-length}, the robot poses adapt to follow the remaining key joints, e.g., ankles and hands. In Fig. \ref{fig:correspondence}, we remove the knee or elbow requirement under scaled link lengths, and the generated motion adapts to better follow the remaining keypoints.

Furthermore, our model can handle unseen heterogeneous robots with relatively low-cost adaptation. We incorporate the skeleton data of Valkyrie and Talos as desired graphs into the original training dataset, then further train the pretrained model for about 5 hours on the same device. The qualitative results are shown in Fig.~\ref{fig:main_result}, and Table~\ref{tab:quant-result} shows comparable performance.


\begin{figure}[t]
    \centering
    \includegraphics[width=0.84\linewidth]{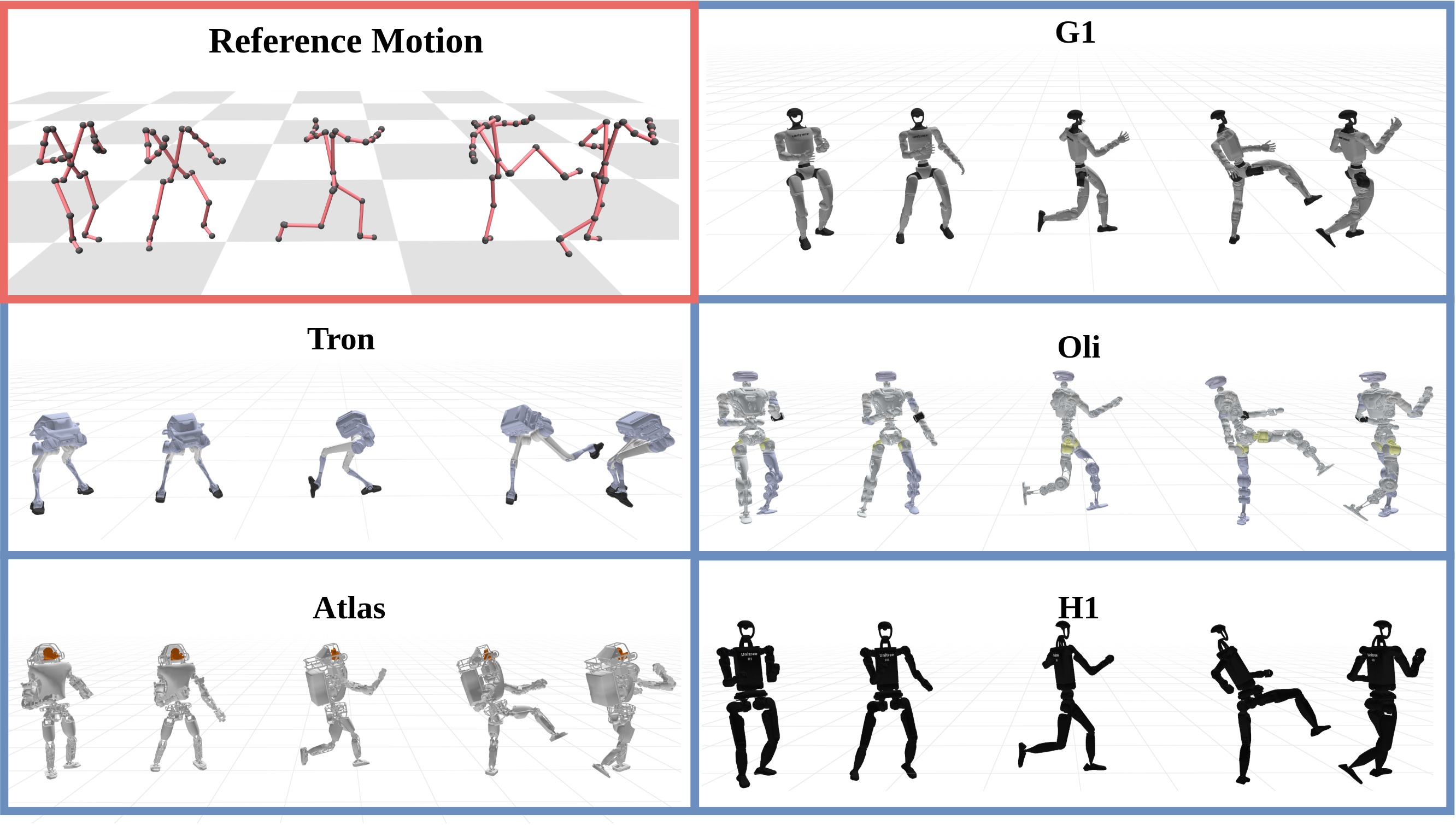}
    \caption{The retargeting results given an out-of-domain reference motion.}
    \label{fig:new_motion}
\end{figure}

Fig.~\ref{fig:new_motion} and the supplementary video show unseen-motion results, and Table~\ref{tab:quant-result} compares in-domain and out-of-domain performance. High-variance failures mainly come from dynamic or recovery-like clips with abrupt root/contact changes, especially on Atlas where morphology amplifies keypoint mismatch without contact, balance, or torque terms.

\subsection{Application}
\begin{figure}[t]
    \centering
    \includegraphics[width=0.90\linewidth]{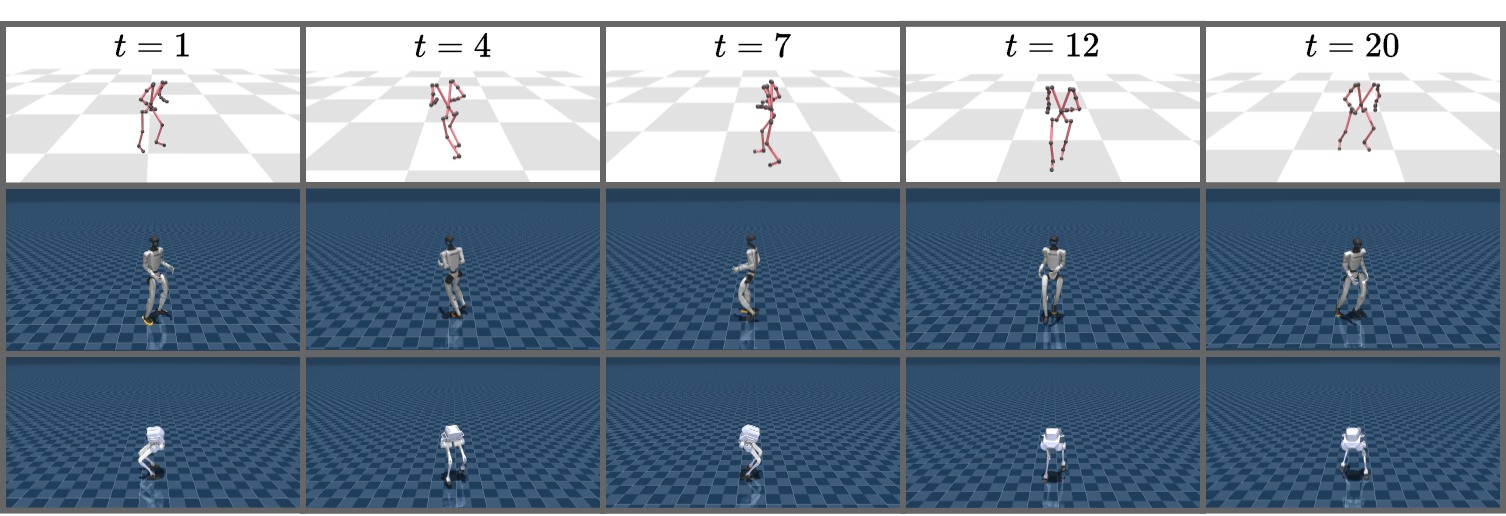}
    \caption{Long-horizon kinematically feasible retargeted motions for whole body tracking task. }
    \label{fig:sim2sim}
\end{figure}
\begin{figure}[htbp]
    \centering
    \includegraphics[width=0.78\linewidth]{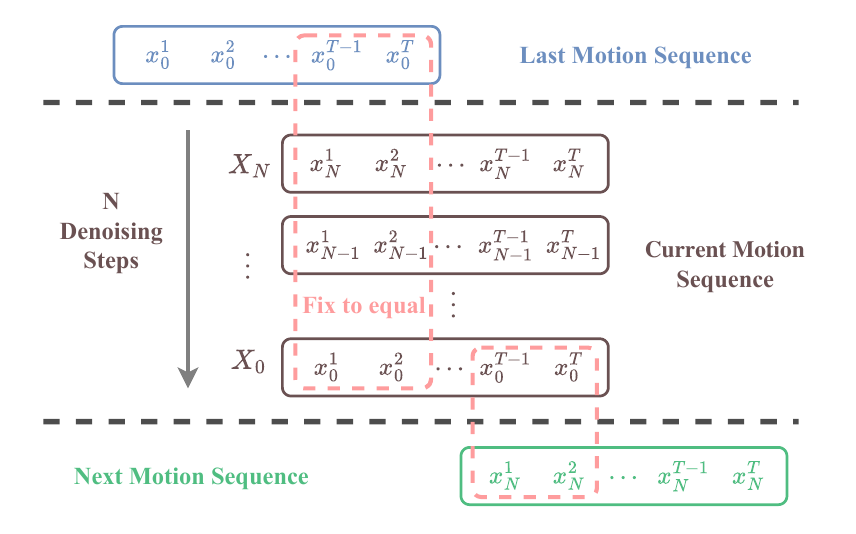}
    \caption{Long-horizon motion inference using diffusion model. }
    \label{fig:long_infer}
\end{figure}
\subsubsection{Long-Horizon Motion Inference}
To use the retargeted motions for imitation tasks, we need references longer than the 60-frame inference horizon. Direct concatenation causes transition discontinuities, so we apply a fixed mask that constrains the first frames of the current clip to match the last frames of the previous one, as shown in Fig.~\ref{fig:long_infer}.

\subsubsection{Sim2Sim Results}
Given the corresponding long-horizon kinematically feasible motions for different embodiments, we choose the humanoid G1 and the biped Tron1 to execute the motion imitation tasks. In detail, we train motion tracking policies using the BeyondMimic~\cite{liao25} pipeline without state estimation module for these two robots. To emphasize the scenario our method is designed to handle, we selected a 20-second reference motion for these two non-homeomorphic robots to imitate. As shown in Fig.~\ref{fig:sim2sim} and in our supplementary video, they can successfully perform the same motions in the physical simulation environment. These results validate the effectiveness and practicality of our method for downstream tasks.

\subsection{Ablation Study}
Since the training loss relies on skeleton information from multi-conditional cross-attention, Table~\ref{tab:abl-result} focuses on spatial and temporal attention. From the comparison results, the framework fails when either component is removed, while temporal attention plays a more critical role in extracting motion features than spatial attention.

\begin{table}[htbp]
\caption{Ablation study on attention blocks.}
\label{tab:abl-result}
\centering
\small
\resizebox{\columnwidth}{!}{%
{\fontsize{8}{8}\selectfont
\begin{tabular}{lllll}
\toprule
\textbf{Model} & $E_{\text{g-mpbpe}} \downarrow$ & $E_{\text{mpboe}} \downarrow$ & $E_{\text{mpbve}} \downarrow$ & $E_{\text{smooth}} \downarrow$ \\
\midrule
\rowcolor[HTML]{EFEFEF}
\multicolumn{1}{c}{\cellcolor[HTML]{EFEFEF}\begin{tabular}[c]{@{}c@{}}w/o spatial attention\end{tabular}} &
   &
   &
   &
   \\
G1            &  0.36\textsubscript{$\pm$0.37}  & 2.03\textsubscript{$\pm$1.03} & 0.33\textsubscript{$\pm$0.30} & 0.043\textsubscript{$\pm$0.013}  \\
Oli           &  0.49\textsubscript{$\pm$0.43}  & 2.11\textsubscript{$\pm$1.16} & 0.58\textsubscript{$\pm$0.44} & 0.041\textsubscript{$\pm$0.014}   \\
Tron1         &  0.27\textsubscript{$\pm$0.24}  & 1.68\textsubscript{$\pm$1.10} & 0.34\textsubscript{$\pm$0.28} & 0.034\textsubscript{$\pm$0.011}   \\
\midrule
\rowcolor[HTML]{EFEFEF}
\multicolumn{1}{c}{\cellcolor[HTML]{EFEFEF}\begin{tabular}[c]{@{}c@{}}w/o temporal attention\end{tabular}} &
   &
   &
   &
   \\
G1            &  1.72\textsubscript{$\pm$2.18}  & 5.64\textsubscript{$\pm$2.30} & 1.54\textsubscript{$\pm$1.80} & 0.062\textsubscript{$\pm$0.013}  \\
Oli           &  2.51\textsubscript{$\pm$3.10}  & 5.76\textsubscript{$\pm$2.50} & 2.72\textsubscript{$\pm$3.11} & 0.069\textsubscript{$\pm$0.028}  \\
Tron1         &  1.94\textsubscript{$\pm$2.24}  & 4.57\textsubscript{$\pm$2.10} & 2.06\textsubscript{$\pm$2.31} & 0.035\textsubscript{$\pm$0.011}  \\
\midrule
\rowcolor[HTML]{EFEFEF}
\multicolumn{1}{c}{\cellcolor[HTML]{EFEFEF}\begin{tabular}[c]{@{}c@{}}ours\end{tabular}} &
   &
   &
   &
   \\
G1            &  \textbf{0.0071\textsubscript{$\pm$0.0038}}  & \underline{0.30\textsubscript{$\pm$0.19}} & \textbf{0.0075\textsubscript{$\pm$0.0035}} & \underline{0.023\textsubscript{$\pm$0.0073}}  \\
Oli           &  \underline{0.011\textsubscript{$\pm$0.0052}} & \textbf{0.25\textsubscript{$\pm$0.17}} & \underline{0.013\textsubscript{$\pm$0.0052}} & \underline{0.023\textsubscript{$\pm$0.0084}} \\
Tron1         &  0.015\textsubscript{$\pm$0.0054} & 0.57\textsubscript{$\pm$0.27} & 0.029\textsubscript{$\pm$0.0088} & \textbf{0.020\textsubscript{$\pm$0.0089}} \\
\bottomrule
\end{tabular}}
}
\end{table}

\section{Conclusion}
\label{sec:conclusion}

In this work, we present a unified transformer-based diffusion framework for multi-embodiment robotic motion generation with retargeting guidance from a reference motion. By representing heterogeneous robotic skeletons with graph structure and incorporating customized attention mechanisms, our model effectively captures topological and kinematic differences across different robots. The proposed energy-based retargeting guidance further enables training without ground-truth target-motion supervision, alleviating the scarcity of motion datasets for diverse robotic embodiments. Extensive experiments across multiple robotic platforms demonstrate that our framework can generate coherent and feasible motions in a unified manner, while exhibiting both skeleton-level and motion-level generalization. The ablation studies analyze the effectiveness of each component in our framework. Moreover, the successful deployment of long-horizon generated motions in a whole-body tracking application highlights the practicality of the proposed approach.

This work offers a promising step toward scalable multi-embodiment robotic motion generation. In the future, we will extend our framework by replacing the reference-motion condition with richer multimodal inputs, such as vision or language. In addition, the dynamics information can be integrated into our framework to generate dynamically feasible motions that can be directly executed on real robotic systems.





\FloatBarrier
\begingroup
\footnotesize
\bibliographystyle{IEEEtran}
\bibliography{ref}  
\endgroup

\end{document}